\documentclass[pmlr]{jmlr}

\usepackage{longtable}% for long tables

\usepackage{booktabs}
\usepackage[load-configurations=version-1]{siunitx} % newer version
\usepackage{appendix}
\theorembodyfont{\upshape}
\theoremheaderfont{\scshape}
\theorempostheader{:}
\theoremsep{\newline}

\title[Automatic Scoring of Students’ Science Writing Using Hybrid Neural Network]{Automatic Scoring of Students’ Science Writing Using Hybrid Neural Network}

 \author{\Name{Ehsan Latif} \Email{ehsan.latif@uga.edu}\\
  \Name{Xiaoming Zhai} \footnote{Corresponding Author: Xiaoming Zhai, 125M Aderhold Hall, 110 Carlton St., Athens, GA 30602}
  \Email{xiaoming.zhai@uga.edu}\\
  \addr AI4STEM Education Center, University of Georgia, Athens, GA, USA}

\begin{document}

\maketitle

\begin{abstract}
This study explores the efficacy of a multi-perspective hybrid neural network (HNN) for scoring student responses in science education with an analytic rubric. 
%. Prior machine learning-based automatic scoring approaches relied on standard specialized models with the limitations of binary assessment questions in capturing cognitive engagement and application of knowledge.  We aim to address these limitations using
%a  HNN model for label semantics and fine-tuned text data for automatic scoring . We have performed experiments on more than 1000 student responses and validated the accuracy with human-rated responses. 
We compared the accuracy of HNN model with four ML approaches (BERT, ANN, Naive Bayes, and Logistic Regression). The results have shown that HHN achieved 8\%, 3\%, 1\%, and 0.12\% higher accuracy than Naive Bayes, Logistic Regression, ANN, and BERT, respectively, for five scoring aspects ($p<0.001$). The overall HNN's perceived accuracy (M = 96.23\%, SD = 1.45\%) is comparable to the (training and inference) expensive BERT model's accuracy (M = 96.12\%, SD = 1.52\%). We also have observed that HNN is x2 more efficient in terms of training and inferencing than BERT and has comparable efficiency to the lightweight but less accurate Naive Bayes model. Our study confirmed the accuracy and efficiency of using HNN for automatically scoring students' science writing.
\end{abstract}

\section{Introduction}
\label{sec:intro}
The shift towards using constructed response questions in science education has been substantial in recent years. Unlike binary assessment questions, such as multiple-choice, constructed responses offer a more robust platform for students to demonstrate their understanding and application of disciplinary core ideas and crosscutting concepts in science. Constructed response questions in science education are essential for eliciting students' cognitive engagement and their ability to apply disciplinary core ideas and crosscutting concepts, as emphasized in the \textit{Framework for K-12 Science Education} \citep{national2012framework}. These questions facilitate the development of emergent and innovative ideas, contributing to creativity and critical thinking. However, evaluating these responses is time-consuming, highlighting the need for efficient automated scoring systems \citep{national2014developing, Nehm2012-aj, zhai2021practice}.

This paper presents an in-depth study focused on implementing and assessing a Hybrid Neural Network (HNN) for the automatic scoring of students' written responses in science education. The overarching objective is to address the limitations inherent in current machine learning (ML) approaches to scoring and to compare the performance of HNN against established ML models. This investigation is structured around two core research questions:

\begin{enumerate}
    \item How does the scoring accuracy of written scientific explanations by the HNN model compare to other models such as BERT, Bayesian, regression, and ensemble algorithms?
    \item What are the differences in training and inferencing efficiency between the HNN model and other ML models, including multi- and individual-perspective BERT scoring algorithms?
\end{enumerate}

The importance of constructed response questions in science education is well-documented, particularly for their role in encouraging cognitive engagement and the application of core scientific principles \citep{national2012framework}. These types of questions are instrumental in fostering students' creative and critical thinking skills by allowing them to explore and articulate emergent ideas in the context of scientific practices \citep{national2014developing}. However, the evaluation of such responses can be labor-intensive and time-consuming, creating a need for effective and efficient automated scoring systems \citep{Nehm2012-aj, zhai2021practice}. Despite significant advancements in the field of ML, including the adoption of deep learning techniques like ANNs, many existing scoring models remain rooted in older, less sophisticated algorithms \citep{zhai2020applying, Platt1998-ky, Boser1992-ih}. Recent developments, such as the application of BERT and SciEdBERT models, have shown promise in enhancing the accuracy of scoring student responses \citep{Riordan2020-pt, Liu2023-pq}.

Despite over a decade of research in ML algorithms for scoring student responses \citep{Nehm2012-ia}, many studies have relied on outdated models, lacking the advanced capabilities of modern deep learning \citep{zhai2020applying, Platt1998-ky, Boser1992-ih}. Recent advancements in artificial neural networks (ANNs) and deep learning, like BERT and SciEdBERT, show promise for accurate scoring \citep{Riordan2020-pt, Liu2023-pq}. Furthermore, fine-tuning Large Language Models (LLM) such as GPT-3.5-Turbo \citep{latif2023fine} and applying chain-of-thought over ChatGPT \citep{lee2023applying} has also been used in literature for automatic scoring. However, given the context-dependent nature of automatic scoring, there is a need for empirical evidence to establish the superiority of these algorithms.

Our study introduces a novel approach using a multi-perspective HNN to address these challenges. This model is designed to leverage the nuanced capabilities of deep learning, focusing on label semantics and fine-grained text data to achieve more accurate and efficient scoring of student responses. The HNN approach is evaluated against four prevalent ML models: BERT, ANN, Naive Bayes, and Logistic Regression. Our methodology involves a comprehensive analysis of over 1000 student responses, focusing on comparing the accuracy and efficiency of these models in real-world educational settings.

The significance of this study lies in its potential to revolutionize the way student responses in science are scored. By providing empirical evidence of the efficacy of HNN in comparison to other ML models, this research contributes to the ongoing dialogue in educational technology about the most effective ways to assess and support student learning in science. The findings have implications not only for the field of educational assessment but also for the broader application of AI in education \citep{latif2023ai}, aligning with the growing interest in Artificial General Intelligence and its potential in educational contexts \citep{latif2023artificial, lee2023multimodality}.

Our primary objectives include demonstrating the enhanced accuracy and efficiency of HNN in scoring student responses, establishing the HNN model as a viable alternative to traditional ML approaches, and contributing to the development of more effective tools for educational assessment.

\section{Literature Review on Machine Learning-based Science Assessment }
Researchers have started utilizing machine algorithms to automatically score students’ science learning for over a decade \citep{RN3402}. The early uses were primarily focused on scoring students’ understanding of scientific concepts. For example, \citet{Nehm2012-aj} used a Summarization Integrated Development Environment (SIDE) program to automatically score students’ written responses to evaluate students' understanding of evolution. SIDE provided various algorithms for users to select based on their data. Specifically, Nehm and his colleagues examined the capabilities of the SMO-type algorithm to score students’ explanations of evolution with a simple rubric that looked at three aspects of the student responses and delivered binary scores for each element. In this early study, the authors found high accuracy and agreement between human and machine scores (Cohen’s kappa $> 0.80$). However, this study noted that the ML performed robustly at the individual item level; it could not deliver similar results when groups of items or more complex rubrics were applied,  \citep{Nehm2012-ia}. 

Given the popularity of SIDE, the creators at Carnegie Melon University updated the program to LightSIDE by including more user-friendly features, such as extracting textual features. In another example, \citet{Nakamura2016-rn} used LightSIDE to automatically score student responses to nine physics scenarios with high school students. The algorithm gave each scenario a binary label similar to the \citet{Nehm2012-ia}. However, this study found only a moderate agreement between human and machine scores despite the more advanced algorithm (Cohen’s kappa 0.33-0.67). Researchers noted that one possible reason for the discrepancy between the two studies could be their relatively little training data set than that in the \citet{Nehm2012-ia} study, showing the complexity of training ML algorithms for this purpose \citep{Nakamura2016-rn}. Besides the open sources such as LightSIDE, \citet{Liu2016-ec} employed a c-rater scoring system to evaluate students’ written responses about knowledge integration (KI) in solving scientific problems.  The c-rater system uses support vector recognition (SVR), an advance from the earlier SVM described in \citep{drucker1996support}, and achieves 0.62-0.90 accuracy. 

With the development of deep learning, \citet{Riordan2020-pt} utilizes ANN to score students’ written responses. \citet{Riordan2020-pt}  used a form of ANN--BERT for natural language processing \citep{Devlin2018-kg}. Researchers compared the human-machine scoring agreement of over two previous ML (SVR and an essential ANN), and found that BERT yielded significantly higher agreement than the earlier models. However, these results showed Cohen’s kappa results between 0.41-0.84, implying a significant variability between tasks and inconsistent scoring results \citet{Riordan2020-pt}. The diverged accuracy among studies may be due to the complexity of construct \cite{haudek2021exploring}, yet \cite{Zhai2021-aa} found algorithms and scoring rubrics can significantly impact accuracy.

\section{Hybrid Neural Networks}
Despite their superiors, different types of ANNs have varying strengths and weaknesses. For example, convolutional neural networks (CNNs) are good at handling image data, while recurrent neural networks (RNNs) handle sequential data such as natural language text. By combining these ANNs with symbolic or knowledge representation using word embedding in a hybrid architecture, we create a better system--HNNs, to handle complex data inputs, generate accurate predictions or classifications, and leverage the strengths of different neural networks (see \ref{fig:architecture} for details of HNNs).

In natural language processing, HNNs involve both the neural network properties seen in deep learning algorithms, such as the transformer-based algorithms discussed above, and symbolic approaches, such as Label Graph Embedding \citep{ma2022hybrid} within the algorithm to allow a more human-like neural network and decision-making process \citep{tahmasebi2012hybrid}. Computer scientists create the HNN algorithm based on the question or problem to be addressed and the type of optimization required. HNN has been used in education to analyze complex problems with multiple inputs that might otherwise be challenging for traditional models. Predicting students' educational performance given more than 15 inputs is challenging, even using existing ANNs \citep{Yousafzai2021-xp}. For example, in one study, researchers looked at the issue of automated essay grading by ML systems and the difficulty in simultaneously evaluating linguistic, structural, and semantic aspects of the wholistic essay. In this case, using an HNN resulted in a 1.4\% increase in Cohen’s kappa over other deep learning approaches \citep{Li2022-sm}. 

Though ML has shown promise to score multi-perspective learning automatically, existing research can only score students’ responses per a single set of labels \citep{Maestrales2021-ne}. If one intends to use multiple sets of labels to assess students’ performance on the task, they have to develop multiple algorithmic models. Being able to score multi-dimensional assessments is challenging as researchers noted that as the complexity of the argument increased, specifically when the algorithm was looking for multiple dimensions, scoring accuracy and agreement decreased \citep{jescovitch2021comparison}. To further the research, a multi-label ML-based text classification approach can apply labels to all aspects simultaneously rather than scoring each element of multiple aspects.

\section{Methodology}
This study assessed middle school student's ability to use energy to explain scientific phenomena. Aiming to support students in developing knowledge-in-use across grade levels, the Next Generation Science Standards (NGSS) provide performance expectations for K–12 students that integrate disciplinary core ideas (DCIs), crosscutting concepts (CCCs), and science and engineering practices (SEPs). The task employed in this study is aligned with the NGSS performance expectation at the middle school level: \textit{Students analyze and interpret data to determine whether substances are the same based on characteristic properties} \citep{national2013next}. This performance expectation requires students to be able to use the structure and properties of matter and chemical reactions (DCIs) and patterns (CCC) to analyze and interpret data (SEP).

\textbf{Participants.} More than 1000 students from grades 6-8 participated in this study. Middle school teachers in the US were offered the opportunity for their students to test open-ended NGSS-aligned science tasks \citep{Zhai2022-cs}.

\begin{figure}
    \centering
    \includegraphics[width=\linewidth]{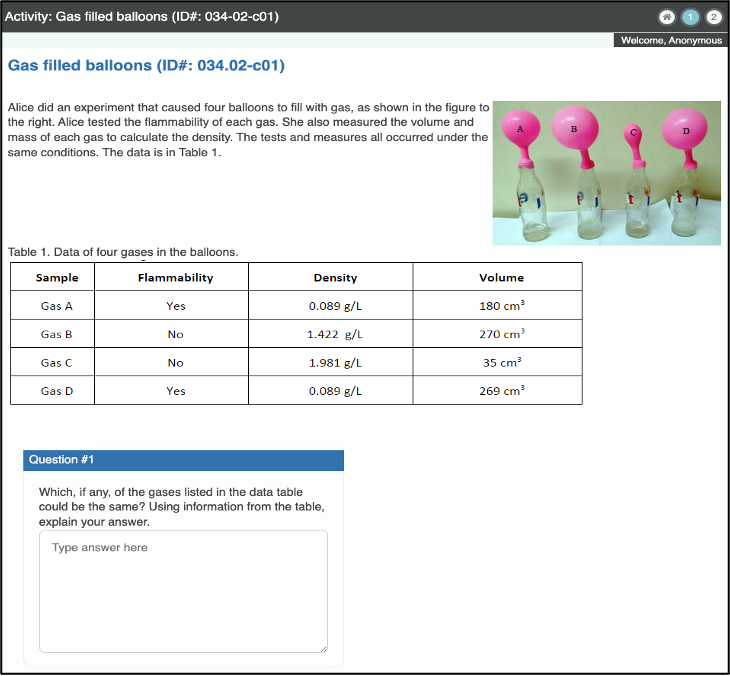}
    \caption{Assessment Item: Gas Filled Balloon}
    \label{fig:assessment_item}
\end{figure}

\textbf{Assessment Item and Human Scoring.} One assessment task was selected from the Next Generation Science Assessment \citep{Harris2024Creating}. This question dealt with the application of chemistry ideas in real-world situations. This item comes under the physical sciences category of "Matter and its Characteristics," which assesses students' analytical and interpreting skills to compare different substances based on their distinctive properties (See Fig.~\ref{fig:assessment_item}). We developed a scoring rubric (see Table~\ref{tab:rubric}) containing five response perspectives associated with each scoring aspect based on the dimensions of science learning: SEP+DCI, SEP+CCC, SEP+CCC, DCI, and DCI \citep{He2024G}.

\begin{table}[htp]
\centering
\caption{Scoring rubric for item “Gas-filled balloons.”}
\begin{tabular}{|p{2cm}|l|p{10cm}|}
\hline
\textbf{Scoring Aspect} &\textbf{Perspectives} & \textbf{Elements} \\
\hline
1 & SEP+DCI & Student states that Gas A and D could be the same substance. \\
\hline
2 &SEP+CCC & Student describes the pattern (comparing data in different columns) in the flammability data of Gas A and Gas D, which is the same as in the table. \\
\hline
3 &SEP+CCC & Student describes the pattern (comparing data in different columns) in density data of Gas A and Gas D, which is the same in the table. \\
\hline
4& DCI & Student indicates flammability is one characteristic of identifying substances. \\
\hline
5 & DCI & Student indicates density is one characteristic of identifying substances. \\
\hline
\end{tabular}
\label{tab:rubric}
\end{table}

% \textbf{Human Scoring}
Ten qualified raters, including experienced middle school science instructors, graduate students, and researchers, were chosen as the human scorers. 
%To create the multi-perspective grading rubric, they conducted group training. After receiving initial training, raters were given one of ten randomly selected portions of the student CRs and asked to submit their initial ratings for each aspect. 
Inter-rater reliability was tested until each rater's IRR of $k > 0.70$ was reached. After the practice sets, the raters progressed to bulk score sets of various sizes with excellent agreement. Further, IRR testing was conducted after mass scoring to guarantee uniformity. %Finally, a training set for the ANN algorithm was produced using the human scores.  For the algorithm to succeed, the training set's quality was essential; thus, addressing problems that made human scoring less reliable was crucial.

% \subsection{Data Processing}
% Before its use for algorithm training, the human-scored data sets were cleaned for use in training and testing. The researchers removed blank replies, nonsensical responses, one-word answers, and non-answers like "I do not know" from the database. One thousand two hundred fifty replies were obtained as a result of this sample cleaning. The answers' beginnings were cleaned up by removing spaces, punctuation, symbols, and digits for the machine-learning algorithms to read the files. The sample was then divided for the algorithms into two or three files. 
For the non-deep learning methods, 20\% of the responses went randomly into the testing set, and 80\% went into the training set. For deep learning algorithms, 60\% of the data was given to the training set, 15\% to the validation set, and 15\% to the testing set.

\textbf{Multi-Perspective HNN for automatic scoring.}
We utilized an HNN for multi-perspective automatic scoring of science responses. The HNN involves BERT for word embeddings, a Bi-LSTM layer for sequential processing, and an attention mechanism for weighting the importance of different input parts.

\begin{figure}
    \centering
    \includegraphics[width=\linewidth]{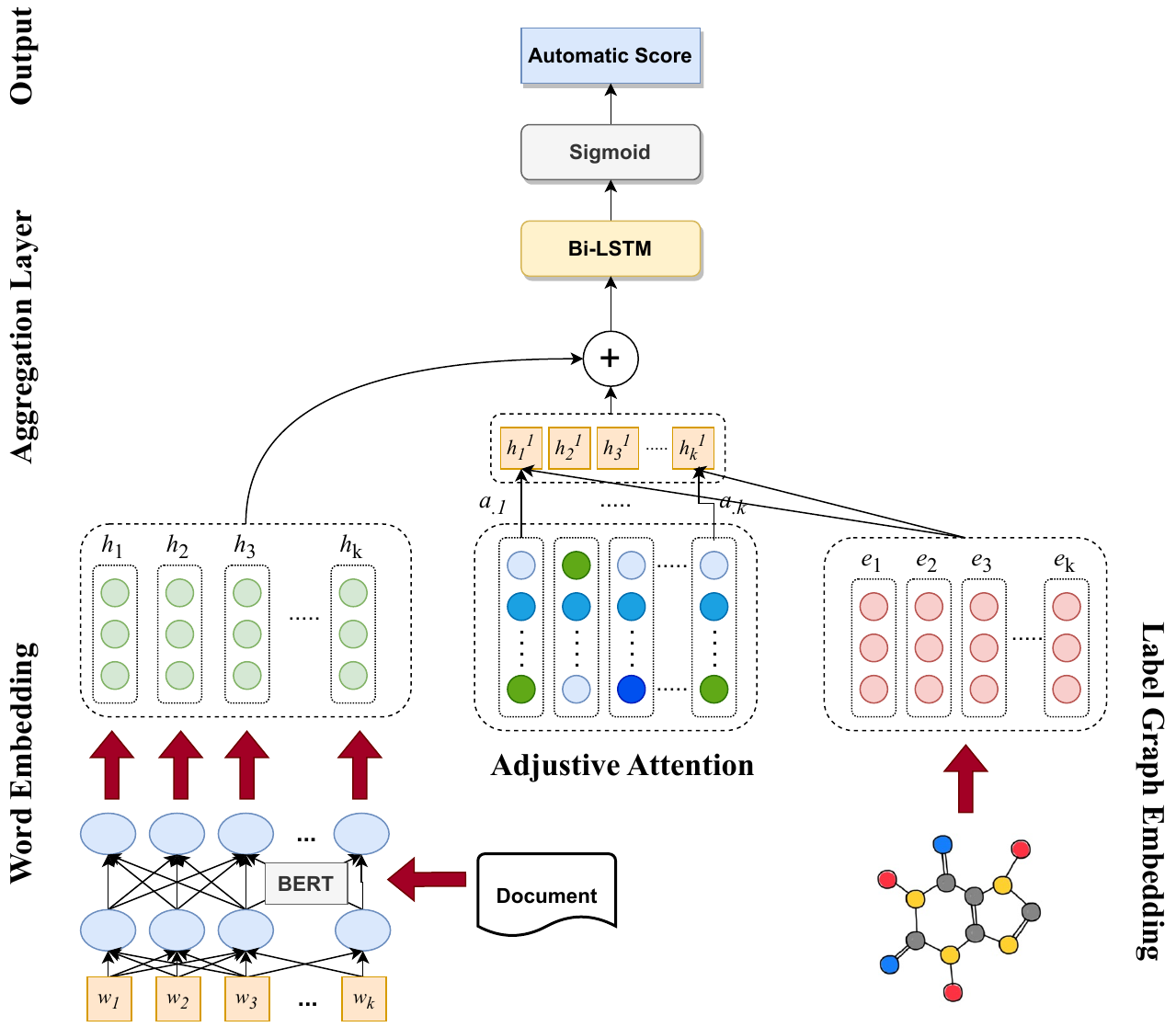}
    \caption{Hybrid Neural Network Architectrue}
    \label{fig:architecture}
\end{figure}

Here is a description of HNN architecture presented in Fig.~\ref{fig:architecture}:

\textbf{Input Layer:} A document (student response) is inputted into the system, where the text is tokenized into words $w_1,w_2,w_3,...w_k$. The student responses, on average, are 150 characters long; hence, the input is capped at 150 bytes.

\textbf{Word Embedding Layer:} Each tokenized word is passed through a BERT model to generate embeddings $h_1,h_2,h_3,...h_k$, represented by circles connected in a horizontal layout, each with a unique color.

\textbf{Sequential Processing Layer:} The word embeddings feed into a Bi-LSTM layer, depicted as two rows of interconnected nodes, indicating the forward and backward passes, with arrows showing the direction of data flow.

\textbf{Attention Mechanism:} The outputs of the Bi-LSTM are then passed through an attention mechanism, visualized by a grid of nodes where each node is connected to all Bi-LSTM nodes, with varying thickness of lines representing the attention weights.

\textbf{Aggregation Layer:} The attended features are aggregated, possibly shown as a funnel leading to a single point, emphasizing the consolidation of information.

\textbf{Output Layer:} Finally, the aggregated information is passed through a sigmoid activation function, resulting in the final output (automatic score), which could be visualized as a bar chart to signify the prediction probabilities. Based on each perspective mentioned in Table~\ref{tab:rubric}, model assigned score; for example, if a student response contains SEP+DCI, its score will be 1.

\textbf{Model Training.}We have trained four machine learning models: ANN \citep{ghiassi2012automated}, BERT \citep{Devlin2018-kg}, Naive Bayes \citep{kolluri2020withdrawn} and Logistic Regression \citep{genkin2007large} using the dataset collected from student responses. LightSIDE \footnote{Carnegie Mellon University’s Language Technologies Institute; http://ankara.lti.cs.cmu.edu/side/} was utilized for the ANN, Naive Bayes, and Linear Regression algorithm training. %During the pre-training phase, the model is trained on a sizable corpus of text data, which aids in teaching it a general understanding of the language. 
The BERT training process was performed using fine-tuning with our data. Our approach trained the model using the written script in Python for 50 epochs to ensure high validation accuracy. Standard program settings of 10 random folds for cross-validation were utilized in both cases for consistency. 
%As described above, the data set was divided into 80\% training (1000 human-labeled student responses) and 20\% testing (250 masked-human-labeled student responses). Following the training, the accuracy rate of the training for each algorithm was recorded. 

\section{Results}
Scoring accuracies for the five algorithms on the five scoring aspects as mentioned in Table~\ref{tab:rubric} are shown in Table~\ref{tab:training_accuracy}. Except for Naive Bayes, all five algorithms yielded average scoring accuracy for the five aspects higher than .90. To examine the difference between HNN with other algorithms, we conducted a paired-T test between the multi-perspective HNN and other four conventional ML models: ANN, BERT, Logistic Regression, and Naive Bayes. Given the small sample size, we used a criteria $p <  .10$ to reject the null hypothesis in favor of the alternative hypothesis. Results suggest a significant difference between HNN average accuracy (M = 96.23\%, SD = 1.45\%) on the five scoring aspects with ANN (M = 95.17\%, SD = 1.61\%, $ = .058 < .10$), BERT (M = 96.12\%, SD = 1.52\%, $p = .088 < .10$), Naive Bayes' (M = 88.91 \%, SD = 3.24\%, $p = .03 < .10$) and Logistic Regression (M  = 93.79\%, SD = 3.12\%, $p = .02 < .1$). The Naive Bayes model showed significantly lower accuracy (M = 88.94\%, SD = 7.05\%) compared to all other algorithms (all $p < .001$).

\begin{table}[htp]
\centering
\caption{Accuracy of HNN, ANN, BERT, Naive Bayes (NB), and Logistic Regression}
\begin{tabular}{|l|l|l|l|l|l|}
\hline
\textbf{Scoring Aspect} & \textbf{HNN (\%) }& \textbf{ANN (\%)} & \textbf{BERT (\%)}& \textbf{NB (\%)} & \textbf{Logi Reg (\%)}\\
\hline
  1 & \textbf{97.6} & 95.23 & 95.98 & 80.16 & 92.4 \\
\hline
  2 & \textbf{98.1} & 93.14 & 96.77 & 85.48 & 91.93 \\
\hline
  3 & \textbf{94.8} & 94.6 & 95.54 & 86.63 & 93.93 \\
\hline
  4 & \textbf{97.4} & 96.82 & 95.54 & 95.54 & 95.14 \\
\hline
  5 & \textbf{96.8} & 96.07 & 96.76 & 96.76 & 95.55 \\
\hline
\end{tabular}
\label{tab:training_accuracy}
\end{table}

 \section{Training and Inference Efficiency}
% Many applications of ML require scoring systems, aiming to represent the applicability or relevance for a specific activity. Therefore, these algorithms' training and inference processes are crucial factors in their design. 
In this work, we compared the training and inference efficiency of our multi-perspective scoring algorithm to that of individual-perspective scoring algorithms. We define the training and inference efficiency of the algorithm in terms of the time taken to train an ML model for a given dataset and the time to get scores with high confidence, respectively. Individual-perspective scoring algorithms frequently score the input using just one strategy or model. BERT is a well-known illustration among these algorithms. To score the input, it creates contextualized word embeddings using a deep neural network. Although BERT is a pre-trained algorithm, its numerous parameters make training and inference time-consuming. This makes it less effective than more straightforward algorithms, like Naive Bayes or Logistic Regression, which can be trained and inferred more quickly. The training time of BERT is three times higher than fundamental algorithms (Naive Bayes and Logistic), but inference time is similar to other approaches for a single GPU system.

A simple technique called Naive Bayes determines the likelihood that input will fall into a specific category based on the frequency of words or other features that belong to that category. It is a straightforward method that is effective in classifying texts. However, it is constrained when capturing more intricate correlations between features. Another well-liked text classification approach is logistic regression. It can be quickly learned and inferred, and it is reasonably effective. ANN illustrates an individual-perspective scoring algorithm. Using an ensemble approach, eight algorithms were trained simultaneously, and eight were received based on their performance on a specific task.  It took way more time to train compared to the individual algorithm. However, it might function less effectively in some situations as more sophisticated models like BERT. 

Multi-perspective scoring algorithms, on the other hand, use several methods or models to produce scores. This algorithm can produce more reliable and accurate scores by overcoming the shortcomings of individual-perspective algorithms. For instance, our multi-perspective scoring algorithm combines BERT, Naive Bayes, and rule-based methods to provide scores. As a result, we can accurately and consistently award scores by capturing both the semantic and syntactic aspects of the input. The multi-perspective scoring algorithm takes training time comparable to BERT, but due to the single scoring model, it takes four times less inferencing time than BERT. As inferencing efficiency matters the most for real-time scoring portals, multi-perspective models are more efficient and can provide scoring in less time than other approaches.

\section{Discussion and Conclusions}
Our findings suggest that HNN is more accurate and effective than other single-perspective models, which is consistent with prior findings. For instance,  \citet{minaee2021deep} discovered that multi-perspective deep neural networks perform better for text categorization tasks than single-perspective models. Moreover, the authors reported that combining features can increase the classification task's accuracy. Furthermore, research has shown that multi-perspective models benefit various NLP tasks. For example, \citet{Shen2020-qb} created a multi-perspective neural network approach for sentiment analysis, which outperformed conventional methods. Also, this study discovered that when numerous scoring aspects were taken into account, the performance of their model increased.

% The discussion could also explore possible explanations for the enhanced performance of the multi-perspective deep neural network technique. For example, according to earlier research, employing several features to capture various data angles can help you extract more information and increase model accuracy \citep{Xu2021-vb}. Also, the multi-perspective approach's combination of simple and complex strategies helps balance accuracy and effectiveness \citep{minaee2021deep}.
% Overall, the initial investigation results are consistent with earlier work showing how multi-perspective deep neural network methods benefit text categorization tasks. The discussion emphasizes various explanations for enhanced performance. It proposes that the choice of individual-perspective or multi-perspective algorithms should depend on the particular requirements of the work at hand. The performance of the models on other datasets and classification tasks could be examined in further study to expand on these findings.

In conclusion, the suggested multi-perspective hybrid neural network approach offers a potentially effective means of addressing the difficulties associated with evaluating complex constructs, such as the knowledge-in-use skills required for scientific practice as described in the \textit{Framework for K–12 Science Education}. The results of the current study, which assessed the suggested approach's effectiveness and put it up against other conventional ML models, showed that the multi-perspective approach was more accurate in determining the different perceptual expectations of NGSS tasks. Furthermore, by highlighting the possibilities of using label semantics and fine-grained text data to create a categorized scoring scheme that narrates a student's cognitive process for a particular assignment, this study contributes to the field of automated assessment in education. Future studies will examine how well the suggested strategy may be applied in other educational settings and further develop the methodology to improve its real-world applications.
\vspace{1cm}

\acks{This study is supported by the National Science Foundation (grants numbers 2101104, 2100964, 2101166, \& 2101112). The authors are grateful for the PASTA team members. Any opinions, findings, conclusions, or recommendations expressed in this material are those of the author(s) and do not necessarily reflect the views of the NSF.}

\bibliography{pmlr-sample}

\begin{thebibliography}{37}
\providecommand{\natexlab}[1]{#1}
\providecommand{\url}[1]{\texttt{#1}}
\expandafter\ifx\csname urlstyle\endcsname\relax
  \providecommand{\doi}[1]{doi: #1}\else
  \providecommand{\doi}{doi: \begingroup \urlstyle{rm}\Url}\fi

\bibitem[Boser et~al.(1992)Boser, Guyon, and Vapnik]{Boser1992-ih}
Bernhard~E Boser, Isabelle~M Guyon, and Vladimir~N Vapnik.
\newblock A training algorithm for optimal margin classifiers.
\newblock In \emph{Proceedings of the fifth annual workshop on Computational learning theory}, New York, NY, USA, July 1992. ACM.

\bibitem[Council et~al.(2012)]{national2012framework}
National~Research Council et~al.
\newblock \emph{A framework for K-12 science education: Practices, crosscutting concepts, and core ideas}.
\newblock National Academies Press, 2012.

\bibitem[Council et~al.(2013)]{national2013next}
National~Research Council et~al.
\newblock Next generation science standards: For states, by states.
\newblock 2013.

\bibitem[Council et~al.(2014)]{national2014developing}
National~Research Council et~al.
\newblock Developing assessments for the next generation science standards.
\newblock 2014.

\bibitem[Devlin et~al.(2018)Devlin, Chang, Lee, and Toutanova]{Devlin2018-kg}
Jacob Devlin, Ming-Wei Chang, Kenton Lee, and Kristina Toutanova.
\newblock {BERT}: Pre-training of deep bidirectional transformers for language understanding.
\newblock October 2018.

\bibitem[Drucker et~al.(1996)Drucker, Burges, Kaufman, Smola, and Vapnik]{drucker1996support}
Harris Drucker, Christopher~J Burges, Linda Kaufman, Alex Smola, and Vladimir Vapnik.
\newblock Support vector regression machines.
\newblock \emph{Advances in neural information processing systems}, 9, 1996.

\bibitem[Genkin et~al.(2007)Genkin, Lewis, and Madigan]{genkin2007large}
Alexander Genkin, David~D Lewis, and David Madigan.
\newblock Large-scale bayesian logistic regression for text categorization.
\newblock \emph{technometrics}, 49\penalty0 (3):\penalty0 291--304, 2007.

\bibitem[Ghiassi et~al.(2012)Ghiassi, Olschimke, Moon, and Arnaudo]{ghiassi2012automated}
Manoochehr Ghiassi, Michael Olschimke, Brian Moon, and Paul Arnaudo.
\newblock Automated text classification using a dynamic artificial neural network model.
\newblock \emph{Expert Systems with Applications}, 39\penalty0 (12):\penalty0 10967--10976, 2012.

\bibitem[Harris et~al.(2024)Harris, Krajcik, and Pellegrino]{Harris2024Creating}
C.~J. Harris, J.~S. Krajcik, and J.~W. Pellegrino.
\newblock \emph{Creating and using instructionally supportive assessments in NGSS classrooms}.
\newblock NSTA Press, 2024.

\bibitem[Haudek and Zhai(2021)]{haudek2021exploring}
Kevin~C Haudek and Xiaoming Zhai.
\newblock Exploring the effect of assessment construct complexity on machine learning scoring of argumentation.
\newblock In \emph{Annual Conference of National Association of Research in Science Teaching, Florida}, 2021.

\bibitem[He et~al.(2024)He, Shin, Zhai, and Krajcik]{He2024G}
Peng. He, N.~Shin, X.~Zhai, and J.~Krajcik.
\newblock Guiding teacher use of artificial intelligence-based knowledge-in-use assessment to improve instructional decisions: A conceptual framework.
\newblock In Xiaoming Zhai and Joseph Krajcik, editors, \emph{Uses of Artificial Intelligence in STEM Education}, pages xx--xx. Oxford University Press, 2024.

\bibitem[Jescovitch et~al.(2021)Jescovitch, Scott, Cerchiara, Merrill, Urban-Lurain, Doherty, and Haudek]{jescovitch2021comparison}
Lauren~N Jescovitch, Emily~E Scott, Jack~A Cerchiara, John Merrill, Mark Urban-Lurain, Jennifer~H Doherty, and Kevin~C Haudek.
\newblock Comparison of machine learning performance using analytic and holistic coding approaches across constructed response assessments aligned to a science learning progression.
\newblock \emph{Journal of Science Education and Technology}, 30\penalty0 (2):\penalty0 150--167, 2021.

\bibitem[Kolluri and Razia(2020)]{kolluri2020withdrawn}
Johnson Kolluri and Shaik Razia.
\newblock Withdrawn: Text classification using na{\"\i}ve bayes classifier, 2020.

\bibitem[Latif and Zhai(2023)]{latif2023fine}
Ehsan Latif and Xiaoming Zhai.
\newblock Fine-tuning chatgpt for automatic scoring.
\newblock \emph{arXiv preprint arXiv:2310.10072}, 2023.

\bibitem[Latif et~al.(2023{\natexlab{a}})Latif, Mai, Nyaaba, Wu, Liu, Lu, Li, Liu, and Zhai]{latif2023artificial}
Ehsan Latif, Gengchen Mai, Matthew Nyaaba, Xuansheng Wu, Ninghao Liu, Guoyu Lu, Sheng Li, Tianming Liu, and Xiaoming Zhai.
\newblock Artificial general intelligence (agi) for education.
\newblock \emph{arXiv preprint arXiv:2304.12479}, 2023{\natexlab{a}}.

\bibitem[Latif et~al.(2023{\natexlab{b}})Latif, Zhai, and Liu]{latif2023ai}
Ehsan Latif, Xiaoming Zhai, and Lei Liu.
\newblock Ai gender bias, disparities, and fairness: Does training data matter?
\newblock \emph{arXiv preprint arXiv:2312.10833}, 2023{\natexlab{b}}.

\bibitem[Lee et~al.(2023{\natexlab{a}})Lee, Latif, Wu, Liu, and Zhai]{lee2023applying}
Gyeong-Geon Lee, Ehsan Latif, Xuansheng Wu, Ninghao Liu, and Xiaoming Zhai.
\newblock Applying large language models and chain-of-thought for automatic scoring.
\newblock \emph{arXiv preprint arXiv:2312.03748}, 2023{\natexlab{a}}.

\bibitem[Lee et~al.(2023{\natexlab{b}})Lee, Shi, Latif, Gao, Bewersdorf, Nyaaba, Guo, Wu, Liu, Wang, et~al.]{lee2023multimodality}
Gyeong-Geon Lee, Lehong Shi, Ehsan Latif, Yizhu Gao, Arne Bewersdorf, Matthew Nyaaba, Shuchen Guo, Zihao Wu, Zhengliang Liu, Hui Wang, et~al.
\newblock Multimodality of ai for education: Towards artificial general intelligence.
\newblock \emph{arXiv preprint arXiv:2312.06037}, 2023{\natexlab{b}}.

\bibitem[Li et~al.(2022)Li, Yang, Hu, Geng, Lin, and Li]{Li2022-sm}
Xia Li, Huali Yang, Shengze Hu, Jing Geng, Keke Lin, and Yuhai Li.
\newblock Enhanced hybrid neural network for automated essay scoring.
\newblock \emph{Expert Syst.}, 39\penalty0 (10), December 2022.

\bibitem[Liu et~al.(2016)Liu, Rios, Heilman, Gerard, and Linn]{Liu2016-ec}
Ou~Lydia Liu, Joseph~A Rios, Michael Heilman, Libby Gerard, and Marcia~C Linn.
\newblock Validation of automated scoring of science assessments.
\newblock \emph{J. Res. Sci. Teach.}, 53\penalty0 (2):\penalty0 215--233, February 2016.

\bibitem[Liu et~al.(2023)Liu, He, Liu, Liu, and Zhai]{Liu2023-pq}
Zhengliang Liu, Xinyu He, Lei Liu, Tianming Liu, and Xiaoming Zhai.
\newblock Context matters: A strategy to pre-train language model for science education.
\newblock January 2023.

\bibitem[Ma et~al.(2022)Ma, Liu, Zhao, Liang, Zhang, and Jin]{ma2022hybrid}
Yinglong Ma, Xiaofeng Liu, Lijiao Zhao, Yue Liang, Peng Zhang, and Beihong Jin.
\newblock Hybrid embedding-based text representation for hierarchical multi-label text classification.
\newblock \emph{Expert Systems with Applications}, 187:\penalty0 115905, 2022.

\bibitem[Maestrales et~al.(2021)Maestrales, Zhai, Touitou, Baker, Schneider, and Krajcik]{Maestrales2021-ne}
Sarah Maestrales, Xiaoming Zhai, Israel Touitou, Quinton Baker, Barbara Schneider, and Joseph Krajcik.
\newblock Using machine learning to score multi-dimensional assessments of chemistry and physics.
\newblock \emph{J. Sci. Educ. Technol.}, 30\penalty0 (2):\penalty0 239--254, April 2021.

\bibitem[Minaee et~al.(2021)Minaee, Kalchbrenner, Cambria, Nikzad, Chenaghlu, and Gao]{minaee2021deep}
Shervin Minaee, Nal Kalchbrenner, Erik Cambria, Narjes Nikzad, Meysam Chenaghlu, and Jianfeng Gao.
\newblock Deep learning--based text classification: a comprehensive review.
\newblock \emph{ACM computing surveys (CSUR)}, 54\penalty0 (3):\penalty0 1--40, 2021.

\bibitem[Nakamura et~al.(2016)Nakamura, Murphy, Christel, Stevens, and Zollman]{Nakamura2016-rn}
Christopher~M Nakamura, Sytil~K Murphy, Michael~G Christel, Scott~M Stevens, and Dean~A Zollman.
\newblock Automated analysis of short responses in an interactive synthetic tutoring system for introductory physics.
\newblock \emph{Phys. Rev. Phys. Educ. Res.}, 12\penalty0 (1), March 2016.

\bibitem[Nehm and Haertig(2012)]{Nehm2012-aj}
Ross~H Nehm and Hendrik Haertig.
\newblock Human vs. computer diagnosis of students' natural selection knowledge: Testing the efficacy of text analytic software.
\newblock \emph{J. Sci. Educ. Technol.}, 21\penalty0 (1):\penalty0 56--73, February 2012.

\bibitem[Nehm et~al.(2012)Nehm, Ha, and Mayfield]{Nehm2012-ia}
Ross~H Nehm, Minsu Ha, and Elijah Mayfield.
\newblock Transforming biology assessment with machine learning: Automated scoring of written evolutionary explanations.
\newblock \emph{J. Sci. Educ. Technol.}, 21\penalty0 (1):\penalty0 183--196, February 2012.

\bibitem[Platt(1998)]{Platt1998-ky}
John Platt.
\newblock Sequential minimal optimization: A fast algorithm for training support vector machines.
\newblock April 1998.

\bibitem[Riordan et~al.(2020)Riordan, Bichler, Bradford, Chen, Wiley, Gerard, and Linn]{Riordan2020-pt}
B~Riordan, S~Bichler, A~Bradford, J~K Chen, K~Wiley, L~Gerard, and M~C Linn.
\newblock An empirical investigation of neural methods for content scoring of scientific explanations.
\newblock In \emph{Proceedings of the Fifteenth Workshop on Innovative Use of {NLP} for Building Educational Applications}, pages 135--144. 2020.

\bibitem[Shen et~al.(2020)Shen, Ma, Xiang, Lu, Vallejos, Xu, Huang, and Long]{Shen2020-qb}
Jiaxing Shen, Mingyu~Derek Ma, Rong Xiang, Qin Lu, Elvira~Perez Vallejos, Ge~Xu, Chu-Ren Huang, and Yunfei Long.
\newblock Dual memory network model for sentiment analysis of review text.
\newblock \emph{Knowl. Based Syst.}, 188\penalty0 (105004):\penalty0 105004, January 2020.

\bibitem[Tahmasebi and Hezarkhani(2012)]{tahmasebi2012hybrid}
Pejman Tahmasebi and Ardeshir Hezarkhani.
\newblock A hybrid neural networks-fuzzy logic-genetic algorithm for grade estimation.
\newblock \emph{Computers \& geosciences}, 42:\penalty0 18--27, 2012.

\bibitem[Yousafzai et~al.(2021)Yousafzai, Khan, Rahman, Khan, Ullah, Ur~Rehman, Baz, Hamam, and Cheikhrouhou]{Yousafzai2021-xp}
Bashir~Khan Yousafzai, Sher~Afzal Khan, Taj Rahman, Inayat Khan, Inam Ullah, Ateeq Ur~Rehman, Mohammed Baz, Habib Hamam, and Omar Cheikhrouhou.
\newblock Student-performulator: Student academic performance using hybrid deep neural network.
\newblock \emph{Sustainability}, 13\penalty0 (17):\penalty0 9775, August 2021.

\bibitem[Zhai(2021)]{zhai2021practice}
Xiaoming Zhai.
\newblock Practices and theories: How can machine learning assist in innovative assessment practices in science education.
\newblock \emph{Journal of Science Education and Technology}, 30\penalty0 (2):\penalty0 139--149, 2021.

\bibitem[Zhai and Nehm(2023)]{RN3402}
Xiaoming Zhai and Ross Nehm.
\newblock Ai and formative assessment: The train has left the station.
\newblock \emph{Journal of Research in Science Teaching}, 60\penalty0 (6):\penalty0 1390--1398, 2023.
\newblock \doi{DOI: 10.1002/tea.21885}.

\bibitem[Zhai et~al.(2020)Zhai, Yin, Pellegrino, Haudek, and Shi]{zhai2020applying}
Xiaoming Zhai, Yue Yin, James~W Pellegrino, Kevin~C Haudek, and Lehong Shi.
\newblock Applying machine learning in science assessment: a systematic review.
\newblock \emph{Studies in Science Education}, 56\penalty0 (1):\penalty0 111--151, 2020.

\bibitem[Zhai et~al.(2021)Zhai, Shi, and Nehm]{Zhai2021-aa}
Xiaoming Zhai, Lehong Shi, and Ross~H Nehm.
\newblock A meta-analysis of machine learning-based science assessments: Factors impacting machine-human score agreements.
\newblock \emph{J. Sci. Educ. Technol.}, 30\penalty0 (3):\penalty0 361--379, June 2021.

\bibitem[Zhai et~al.(2022)Zhai, He, and Krajcik]{Zhai2022-cs}
Xiaoming Zhai, Peng He, and Joseph Krajcik.
\newblock Applying machine learning to automatically assess scientific models.
\newblock \emph{J. Res. Sci. Teach.}, 59\penalty0 (10):\penalty0 1765--1794, December 2022.

\end{thebibliography}

\end{document}